\pdfoutput=1

\documentclass[11pt]{article}

\usepackage[final]{acl}

\usepackage{times}
\usepackage{latexsym}

\usepackage[T1]{fontenc}

\usepackage[utf8]{inputenc}

\usepackage{microtype}

\usepackage{inconsolata}

\usepackage{tabularx}
\newcolumntype{s}{>{\hsize=.24\hsize}X}
\usepackage{makecell}
\usepackage{threeparttable}
\usepackage{tablefootnote}
\usepackage{caption}
\usepackage{subcaption}
\usepackage{graphicx}
\usepackage{booktabs}
\usepackage{multirow}
\usepackage{adjustbox}
\usepackage{amsmath}
\usepackage{amssymb}
\usepackage[ruled,linesnumbered]{algorithm2e}
\usepackage{wrapfig}
\usepackage[capitalize]{cleveref}
\AtBeginDocument{%
    \crefname{figure}{Figure}{Figures}%
}

\SetKwComment{tcc}{$\triangleright$~}{}
\SetCommentSty{normalfont}
\SetKwInput{KwInput}{Input}
\SetKwInput{KwOutput}{Output}

\title{Reward-based Input Construction for Cross-document Relation Extraction}

\author{Byeonghu Na\textsuperscript{\textmd{1 *}}, Suhyeon Jo\textsuperscript{\textmd{1 *}}, Yeongmin Kim\textsuperscript{\textmd{1}}, Il-Chul Moon\textsuperscript{\textmd{1 2}} \\
        \textsuperscript{\textmd{1}}KAIST,~ \textsuperscript{\textmd{2}}summary.ai \\
        \texttt{\{byeonghu.na,suhyeonjo,alsdudrla10,icmoon\}@kaist.ac.kr}
        }
    
\begin{document}
\maketitle
\def\thefootnote{\textmd{*}}\footnotetext{Equal contribution}
\def\thefootnote{\arabic{footnote}}

\begin{abstract}
Relation extraction (RE) is a fundamental task in natural language processing, aiming to identify relations between target entities in text. While many RE methods are designed for a single sentence or document, cross-document RE has emerged to address relations across multiple long documents. Given the nature of long documents in cross-document RE, extracting document embeddings is challenging due to the length constraints of pre-trained language models. Therefore, we propose REward-based Input Construction (REIC), the first learning-based sentence selector for cross-document RE. REIC extracts sentences based on relational evidence, enabling the RE module to effectively infer relations. Since supervision of evidence sentences is generally unavailable, we train REIC using reinforcement learning with RE prediction scores as rewards. Experimental results demonstrate the superiority of our method over heuristic methods for different RE structures and backbones in cross-document RE. Our code is publicly available at \url{https://github.com/aailabkaist/REIC}.
\end{abstract}

\section{Introduction}
\label{sec:intro}


\begin{figure*}[t]
    \centering
    \includegraphics[width=\linewidth]{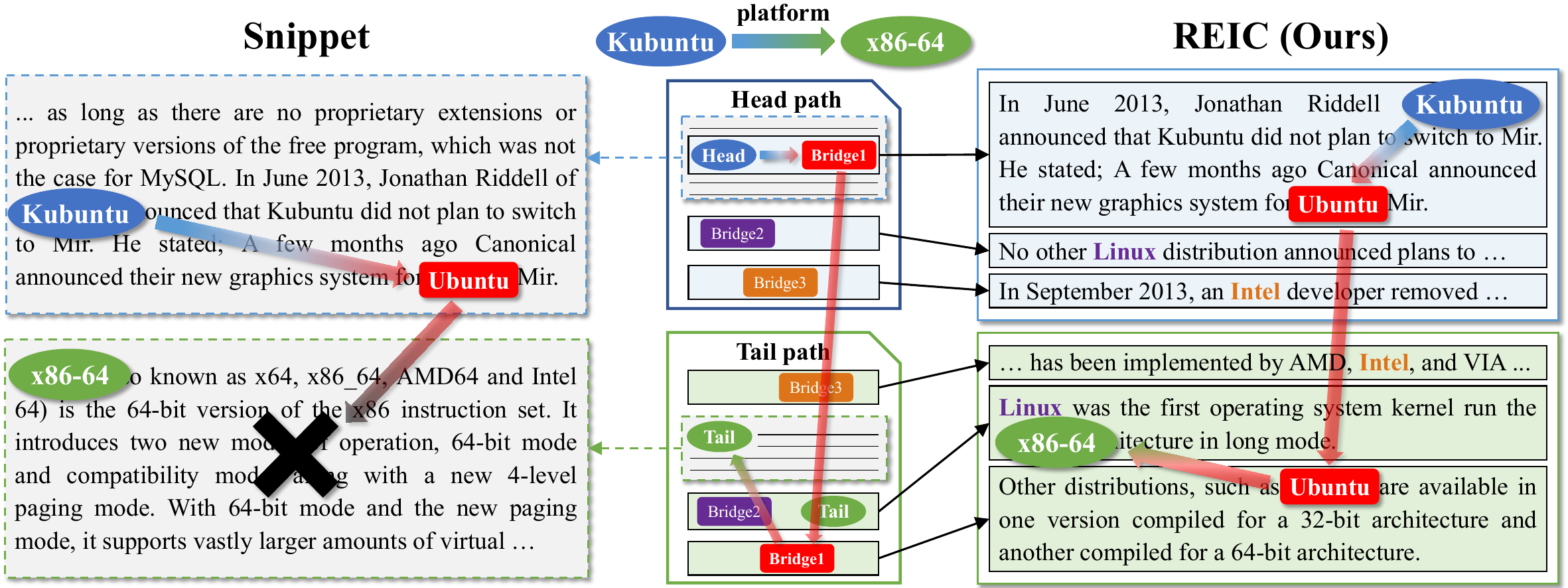}
    \caption{An illustrated comparison between Snippet and selected sentences using our REward-based Input Construction (REIC) for cross-document relation extraction. The figure depicts an example triplet (\texttt{Kubuntu}, \texttt{x86-64}, platform) with the text path (`Mir (software)', `X86-64'), including three bridge entities (\textbf{Ubuntu}, \textbf{Linux}, \textbf{Intel}) abbreviated as (Bridge1, Bridge2, Bridge3). Dash and solid arrows signify the selection process of Snippet and REIC, respectively, while gradient-colored arrows indicate connections between the head and tail entities. REIC selects important sentences from any position within a path to determine the relation between the head and tail entity, whereas Snippet only includes sentences located around the head or tail entity.}
    \label{fig:example}
\end{figure*}

The task of relation extraction (RE) aims to identify relations between a pair of target entities in a given text~\cite{zeng2014relation,zhang2017position, sahu-etal-2019-inter, xu2021entity}. This task is fundamental in natural language processing (NLP) and provides crucial information for applications such as information retrieval~\cite{kadry-open} and question answering~\cite{li-etal-2019-entity,chen-etal-2019-uhop}. Most existing RE methods are limited to scenarios where the entity pair is within a single sentence~\cite{wang2016relation,zhu-etal-2019-graph} or a single document~\cite{nan-etal-2020-reasoning,huang-etal-2021-three,xu2021entity}. However, many scenarios require extracting relations across multiple documents, where the target entity pair may not coexist within the same document. Therefore, recent efforts to address these challenges have led to the proposal of cross-document RE by \citet{yao-etal-2021-codred}, and research in this area has received considerable attention~\cite{wang-etal-2022-entity,wu-etal-2023-local,lu-etal-2023-multi,son-etal-2023-explore}.

Unlike traditional RE research, cross-document RE involves extracting relational facts from large-scale long documents. For example, the DocRED dataset~\cite{yao-etal-2019-docred}, developed for document-level RE, contains an average of 198 words per document. In contrast, the CodRED dataset~\cite{yao-etal-2021-codred}, designed for cross-document RE, has an average of 2,416 words per document. Given the substantial length of documents in cross-document RE tasks, extracting document embeddings from pre-trained language models, a step common to all methods, poses considerable challenges. This is because pre-trained language models are limited by the maximum number of tokens they can process; for example, BERT~\cite{devlin-etal-2019-bert} typically has a limit of 512 tokens.

A simple approach adopted in several studies is to use text snippets around the target entities as input to pre-trained language models, as shown in the dashed boxes in the middle panel of \cref{fig:example}. Often, this text snippet is determined by either word proximity or document structure without any adjustment. However, if only a subset of sentences around the entities is extracted from long documents, there is a risk of missing important sentences that are crucial for RE. For example, as seen in the left panel of \cref{fig:example}, using snippets around the entities fails to capture the relation `platform' between the entity pair (\texttt{Kubuntu}, \texttt{x86-64}).

In this paper, we introduce REward-based Input Construction (REIC), the first learning-based input sentence selection module specifically designed for cross-document RE. The right panel of \cref{fig:example} displays the sentences selected by our model from the example. By identifying relational evidence sentences through the selection module, the RE module infers the relation between the target entities using the reasoning chain illustrated by the gradient-colored arrows in \cref{fig:example}. Our approach is to develop a sentence selection module that computes the probability of selecting sentences based on the currently selected sentences. We show that this sentence selection process can be modeled as a Markov decision process (MDP). We specifically choose to use MDP because there is no supervision for sentence selection from corpus, so RE models have to perform iterative learning by trials. Subsequently, we train the sentence selection module using reinforcement learning (RL), where the relation prediction score obtained from the selected sentences serves as the reward. Through experimental validation, REIC outperforms other heuristic methods across various RE modules.

\section{Related Work}
\label{sec:rel}

\subsection{Cross-document Relation Extraction}

Traditionally, research in RE has primarily focused on the sentence- or document-level, typically dealing with documents no longer than a single paragraph, as reviewed in \cref{sec:app_rel}. However, there has been a growing interest in cross-document RE, which aims to identify relations between entities across multiple long documents. 
\citet{yao-etal-2021-codred} laid the groundwork for cross-document RE by introducing the CodRED benchmark dataset. They also proposed a simple baseline model that utilizes document embeddings obtained from pre-trained language models like BERT~\cite{devlin-etal-2019-bert}. Building on this work, ECRIM~\cite{wang-etal-2022-entity} proposed an entity-based input filtering approach and an entity-based RE module with an attention mechanism, leading to significant performance improvements. Additionally, \citet{wu-etal-2023-local} introduced a local causal estimation algorithm and a globally guided reasoning algorithm based on causality. In addition, \citet{lu-etal-2023-multi} presented a method for multi-hop evidence retrieval through evidence path mining and ranking, and \citet{son-etal-2023-explore} proposed a method for constructing explicit reasoning paths using bridging entities. In our approach, we focus on selecting essential sentences to effectively extract document information from long documents for cross-document RE. Unlike other methods, our approach involves learning-based input construction, and these surveyed models can utilize our input construction method.




\begin{figure*}[t]
    \centering
    \includegraphics[width=\linewidth]{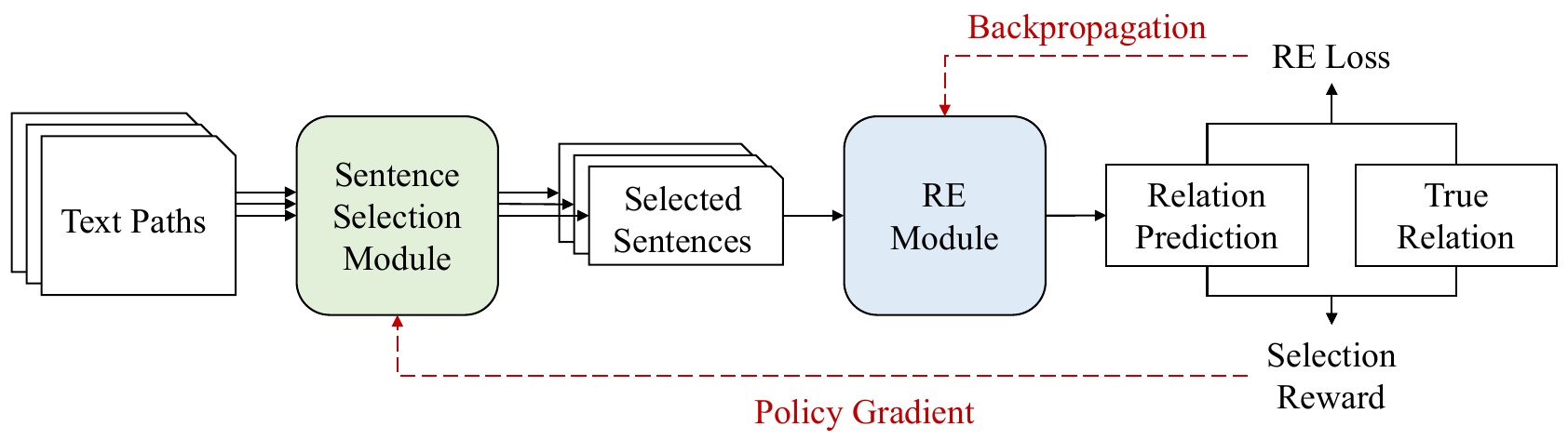}
    \caption{Overview of our cross-document RE framework. Solid black arrows represent forward propagation, while the dashed red arrows represent the loss signal. Documents from each text path are processed by the sentence selection module to construct the input for the RE module, which then obtains relation predictions.}
    \label{fig:overview}
\end{figure*}

\subsection{Sentence Selection in NLP}

Numerous studies have addressed the challenge of selecting relevant sentences from extensive and noisy textual datasets while preserving contextual coherence. These approaches include deep learning-based binary classification for each sentence~\cite{cheng-lapata-2016-neural}, auto-regressive techniques to capture sentence correlations~\cite{zhou-etal-2018-neural-document,zhong-etal-2019-searching,zhang-etal-2023-extractive-summarization}, methods that consider summary-level context~\cite{zhong-etal-2020-extractive}, and the use of diffusion models~\cite{zhang-etal-2023-diffusum}.


Other lines of research utilize RL for sentence selection~\cite{narayan-etal-2018-ranking}. This approach naturally reflects the correlation between sentences, as the selection process is optimized by considering the current state, which includes the sentences already selected. Furthermore, RL offers flexibility in reward shaping, making it suitable for specific downstream tasks. For example, \citet{feng2018reinforcement, qin-etal-2018-robust, qu2019fine,Takanobu_Zhang_Liu_Huang_2019} used RL to filter noisy datasets, thereby improving the performance of sentence-level RE. Also, \citet{xu-etal-2022-evidence} utilized RL to evidence extraction in document-level RE, and \citet{Man_Ngo_Van_Nguyen_2022} applied RL to find input tokens for BERT embedding within documents for event-event RE. Our approach builds on previous methods for sentence selection and represents the first application of RL to cross-document RE.


\section{Methods}
\label{sec:methods}


\subsection{Problem Formulation}
\label{subsec:prob}

The cross-document RE~\cite{yao-etal-2021-codred} aims to infer a relation $r\in\mathcal{R}$ between a pair of target entities $(e^\text{h}, e^\text{t})$, where $e^\text{h}$ is the head entity and $e^\text{t}$ is the tail entity, from a bag of text paths $\{ p_n \}_{n=1}^N$. Each text path $p_n$ consists of two documents $( d_n^\text{h}, d_n^\text{t} )$ that satisfy the following conditions: 1) $d_n^\text{h}$ and $d_n^\text{t}$ contain the head and tail entities, respectively; 2) there exists at least one entity, called a bridge entity, which is shared by both documents.

Our goal is to develop a method that effectively extracts sentences from an input document to facilitate the relation extraction between two entities. Given a document $d$ consisting of $M$ sentences, denoted by $d = \{ s_{m} \}_{m=1}^M$, our approach is to select important sentences and obtain document embeddings using pre-trained language models such as BERT~\cite{devlin-etal-2019-bert}. Given the maximum input length constraint of pre-trained language models, collecting significant sentences enables effective extraction of document information.

\begin{figure*}[t]
    \centering
    \includegraphics[width=\linewidth]{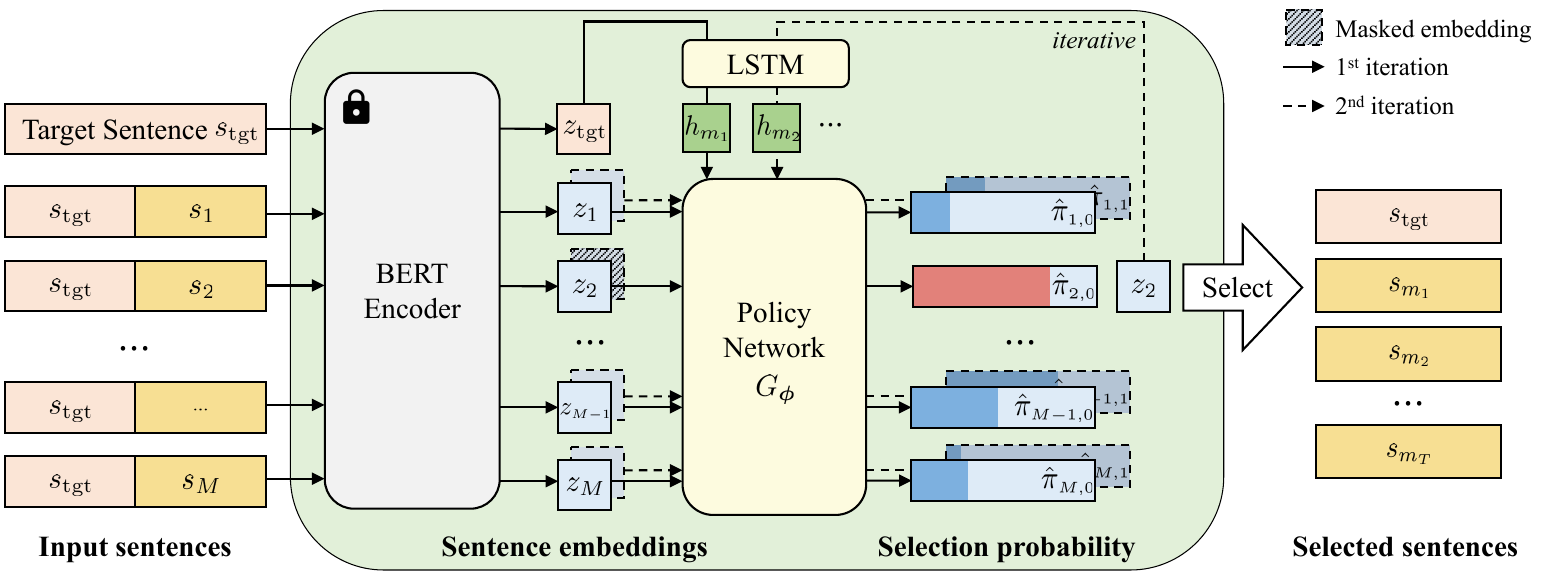}
    \caption{Illustration of the REIC module: BERT encoder, policy network, and LSTM. Each sentence $s_m$, combined with the target sentence $s_{\text{tgt}}$, is processed by the fixed BERT to obtain the embedding $z_m$. Then, these embeddings pass through the policy network $G_{\boldsymbol{\phi}}$ to obtain the selection probability $\hat{\pi}_{m,t}:=\hat{\pi}(s_m|S_t)$, and the sentence is sampled with this probability. The embedding of the selected sentence is fed to the LSTM to incorporate information from previously selected sentences into the policy network. In this way, sentences are sequentially selected, with masking applied to embeddings for subsequent selections, as indicated by the dotted line.}
    \label{fig:method}
\end{figure*}

\subsection{Overall Framework}
\label{subsec:overall}

We begin by describing our proposed framework as depicted in \cref{fig:overview}. We divide the cross-document RE module into the input sentence selection module and the RE module. The goal of the sentence selection module is to identify important sentences for RE in each document. Next, the RE module outputs relation prediction scores based on the sentences identified by the sentence selection module.

Specifically, for each text path $p_n=( d_n^\text{h}, d_n^\text{t} )$, we utilize the sentence selection module to extract a summarized text path $\bar{p}_n=( S_n^\text{h}, S_n^\text{t} )$, containing essential sentences to recognize relations between entity pairs. Afterwards, these summarized text paths become an input to the RE module, which employs a pre-trained language model as its backbone, to produce relation prediction scores $\hat{y}(S_n^\text{h}, S_n^\text{t})$ for each path. The RE module is trained using relation prediction loss. For example, ECRIM~\cite{wang-etal-2022-entity} adopts a multi-label global-threshold loss:
\begin{align}       \label{eq:re_loss}
    \mathcal{L}_{\text{RE}} = & \log (e^{\theta} + \sum\nolimits_{r\in \mathcal{R} \setminus \Omega^{\text{h,t}}} e^{\hat{y}_r} )  \nonumber \\
    & + \log (e^{-\theta} + \sum\nolimits_{r\in \Omega^{\text{h,t}}} e^{-\hat{y}_r} ),
\end{align}
where $\theta$ is the threshold and $\Omega^{\text{h,t}}$ denotes the set of relations that match the relation of the target entity pair. On the other hand, we compute the selection reward to train the sentence selection module using RL, as explained in \cref{subsec:train_ssm}. It is important to note that our sentence selection module can be integrated into any existing RE modules.

\subsection{REIC: Reward-based Input Construction}

We present the proposed sentence selection module, called REward-based Input Construction (REIC), for cross-document RE. We build the structure of the REIC module inspired by \citet{Man_Ngo_Van_Nguyen_2022}, as shown in \cref{fig:method}.
First, we obtain the representation $z_{m}$ of each sentence $s_{m}$. We denote the sentence containing the target entity as $s_{\text{tgt}}$. We are trying to extract information relevant to the target entities in this document. Therefore, it would be helpful to include the information from the sentence that contains the target entity. To achieve this, we concatenate $s_{\text{tgt}}$ and $s_m$ and input them into the pre-trained BERT encoder to generate a sentence embedding $z_m$ for the $m$\textsuperscript{th} sentence:
\begin{align}       \label{eq:senemb}
    z_m = \text{BERT} ([s_{\text{tgt}}, s_m]).
\end{align}
Note that we maintain the positions in the document when concatenating.

Next, we propose an iterative sentence selection process. We introduce the policy network $G_{\boldsymbol{\phi}}$, with trainable parameters $\boldsymbol{\phi}$, for generating the sentence selection probabilities. The term `policy' is used because we treat the sentence selection process as a policy and apply policy gradient methods, as we will explain in \cref{subsec:train_ssm}. The policy network takes the sentence embeddings as input and outputs the probability of selecting each sentence based on the previously selected sentences.

Specifically, we denote the set of selected sentences after iteration $t$ as $S_t$, and we initialize it as $S_0=\{ s_{\text{tgt}} \}$, meaning that the target sentence is selected by default. At iteration $t+1$, we aim to select the next important sentence $s_{m_{t+1}}$ from the given set of $t+1$ selected sentences, $S_{t} = \{ s_{\text{tgt}}, s_{m_1}, ..., s_{m_{t}} \} $. To provide information about the sentences already selected, we introduce the Long Short-Term Memory (LSTM)~\cite{lstm} network to the sentence embedding $z_m$. By feeding the embedding $z_{m_{t}}$ of the selected sentence at time $t$ into the LSTM, the hidden vector $h_{m_{t}}$ of the LSTM serves as the representation of the previously selected sentences:
\begin{align}       \label{eq:lstm}
    h_{m_{t}} = \text{LSTM} (z_{m_t}, h_{m_{t-1}}).
\end{align}
Then, we concatenate the embedding $h_{m_t}$ of the selected sentences with the embedding $z_m$ of each unselected sentence and feed them into the policy network $G_{\boldsymbol{\phi}}$ to obtain the selection probabilities:
\begin{align}       \label{eq:policy_prob}
    \hat{\pi}(s_m | S_{t}) = G_{\boldsymbol{\phi}} ([z_m, h_{m_{t}}]).
\end{align}
We select the sentence from the sampling with the selection probability as the next selected sentence:
\begin{align}       \label{eq:policy_select}
    s_{m_{t+1}} \sim \text{Categorical}(\hat{\pi}(s_m | S_{t})).
\end{align}
This process is iterated for a maximum number of iterations $T$ and is outlined in \cref{alg:selection}.

\begin{algorithm}[t]
    \DontPrintSemicolon
    \KwInput{Input sentences $\{ s_m \}_{i=1}^M$}
    \KwOutput{Selected sentences $S$}
    $\mathcal{M} \leftarrow \{ 1, ..., M \}$ \\
    $z_m \leftarrow \text{BERT}([s_{\text{tgt}}, s_m])$ for $m \in \mathcal{M}$ \\
    $S \leftarrow \{ s_{\text{tgt}} \}$ \\
    $z \leftarrow z_{\text{tgt}} $\\
    $\mathcal{M} \leftarrow \mathcal{M}  \setminus \{ \text{tgt} \}$ \\
    $h \leftarrow \varnothing $ \\
    \For{$t \in \{ 1, ..., T\}$}{%
        $h \leftarrow \text{LSTM} (z, h)$ \\
        $\hat{\pi}_m \leftarrow G_{\boldsymbol{\phi}}([z_m, h])$ for $m \in \mathcal{M}$ \\
        $\hat{m} \sim \text{Categorical}(\hat{\pi}_m) $ \\
        $S \leftarrow S \cup \{ s_{\hat{m}} \}$ \\
        $z \leftarrow z_{\hat{m}}$ \\
        $\mathcal{M} \leftarrow \mathcal{M}  \setminus \{ \hat{m} \}$ \\
    }
    \caption{Algorithm of REIC}     \label{alg:selection}
\end{algorithm}

\subsection{Training REIC with RL}
\label{subsec:train_ssm}

We discuss the method of training the REIC module based on RL. We represent the sentence selection process as a MDP. The state space consists of combinations of selected sentences, which is represented by a set of selected sentences, $S$. Selecting one sentence at each iteration is considered as an action. Thus, when the action of selecting sentence $s_m$ from the current state $S$ is taken, it transitions to the state $S\cup\{s_m\}$. In this case, the policy, which is the distribution of actions given a state, can be represented by the selection probability from \cref{eq:policy_prob}. Once the action is determined; the next state is also determined, so is the transition probability.

The proposed sentence selection module ultimately needs to choose sentences that perform well for the RE task. Therefore, it is necessary to set the reward function to provide signals for RE. An intuitive approach is to identify the important sentences necessary for RE and provide rewards when these sentences are selected. However, obtaining the important sentences from the large dataset requires specialized knowledge and a considerable amount of time, making it extremely challenging. Therefore, we intend to determine the reward as the relation prediction results obtained by performing RE using the selected sentences.

Specifically, for a given text path, considering the selected sentences $S^{\text{h}}$ for the head document and $S^{\text{t}}$ for the tail document, the reward $R$ is,
\begin{align}       \label{eq:reward}
    R(S^{\text{h}}) = R(S^{\text{t}}) = \lambda_r \hat{y}_r(S^{\text{h}}, S^{\text{t}}),
\end{align}
where $\hat{y}_r(S^{\text{h}}, S^{\text{t}})$ is the prediction score of the true relation $r$ obtained by inputting the selected sentence pairs $(S^{\text{h}}, S^{\text{t}})$ into the RE module. Here, $\lambda_r$ is a reward hyperparameter adjusted based on the relation type.
Empirically, due to the dataset imbalance problem where there are more `no relation' (N/A) cases than positive cases, we find that it is beneficial to assign a higher value to positive relations compared to `no relation'. In our experiments, we set $\lambda_r = 10$ for positive relations and $1$ otherwise, as discussed in \cref{subsec:abl}. This hyperparameter $\lambda$ can be considered as a prior of Bayesian RL. Additionally, depending on the loss structure, we also consider replacing it with rewards that are proportional to the prediction scores, which is detailed in \cref{subsec:exp_setting}. 

Based on the defined MDP, we update the policy network directly using one of the policy gradient methods, REINFORCE~\cite{Williams1992}:
\begin{gather}       \label{eq:reinforce}
    \boldsymbol{\phi} \leftarrow \boldsymbol{\phi} + \alpha R(S_T)  \nabla \log \hat{\pi}(S_T), \ \\
    \text{where } \log \hat{\pi}(S_T) = \textstyle \sum_{t=1}^T \log \hat{\pi}(s_{m_t}|S_{t-1}).
\end{gather}
Here, $\alpha$ is the learning rate.
In the implementation, for the sake of time and memory efficiency, we fix the BERT encoder for sentence embedding extraction and only optimize the parameters of the policy network and the LSTM.

We train the REIC module and the RE module simultaneously, as described in \cref{fig:overview}. The REIC module extracts important sentences from documents in a batch, and these selected sentences are used to predict relations through the RE module. Based on the predicted outputs, we compute the reward and the RE loss. Then, we optimize the REIC module and the RE module using policy gradient and backpropagation, respectively.

\begin{table*}[t]
    \centering
    \adjustbox{max width=\linewidth}{%
    \begin{tabular}{ccc | cccc cc}
        \toprule
        \multicolumn{2}{c}{RE module} & \multirow{2}{*}[-0.5\dimexpr \aboverulesep + \belowrulesep + \cmidrulewidth -0.4em]{\shortstack[c]{Input \\construction \\ module}} & \multicolumn{4}{c}{Dev} & \multicolumn{2}{c}{Test} \\
        \cmidrule(lr){1-2} \cmidrule(lr){4-7} \cmidrule(lr){8-9}
         Backbone & Structure  &  & AUC   & F1    & P@500     & P@1000    & AUC   & F1    \\
        \midrule
        BERT & End-to-end & Snippet\textsuperscript{*} & 53.83 & 55.17 & 67.80     & 54.80     & 54.47 & 56.81 \\
        &&ECRIM-IC       & 51.38 & 52.51 & 66.60     & 52.10     & 48.97 & 52.47 \\
        &&REIC (Ours)       & \bf{55.80} & \bf{56.51} & \bf{71.20}     & \bf{56.70}     & \bf{54.71} & \bf{57.38} \\
        \cmidrule(lr){2-9}
        & ECRIM & Snippet          & 57.14 & 57.85 & 76.00     & 57.70     & 58.88 & 60.79 \\
        &&ECRIM-IC       & 62.96 & 61.43 & 78.40     & 61.80     & 62.17 & 60.80 \\
        &&REIC (Ours)         & \bf{64.50} & \bf{63.77} & \bf{78.60}     & \bf{63.80}     & \bf{62.59} & \bf{61.27} \\
        \midrule
        RoBERTa & End-to-end & Snippet  & 56.47 & 57.49 & 72.60     & 57.10     & 56.32 & 56.05 \\
        &&ECRIM-IC     & 54.20 & 55.55 & 69.60     & 55.70     & 54.35	& 54.05\\
        &&REIC (Ours)     & \bf{58.16} & \bf{58.74} & \bf{74.40}     & \bf{58.80}     & \bf{57.71} & \bf{58.35} \\
        \cmidrule(lr){2-9}
        & ECRIM & Snippet   & 63.09 & 63.27 & 80.00     & 63.40     & 64.31     & 64.74 \\
        &&ECRIM-IC        & 59.36 & 61.81 & 79.60     & 61.90     & 62.12     & 63.06 \\
        &&REIC (Ours)       & \bf{66.41} & \bf{63.47} & \bf{80.20}     & \bf{63.50}     & \bf{65.88} & \bf{65.02} \\
        \midrule[\heavyrulewidth]
        \multicolumn{9}{l}{\makecell[l]{\footnotesize * `Snippet + End-to-end' shows a significant performance improvement over the results reported in \citet{yao-etal-2021-codred}. \\[-2pt] ~~~\footnotesize 
  This improvement is due to rectification of minor errors in the original implementation, as detailed in \cref{subsubsec:app_imp_baseline}.}}\\
    \end{tabular}
    }
    \caption{Performance comparison with input construction methods on the development (Dev) and test datasets of CodRED.  Test set results are obtained from the official CodRED website on CodaLab.}
    \label{tab:main}
\end{table*}


\section{Experiments}
\label{sec:exp}

\subsection{Experimental Settings}
\label{subsec:exp_setting}

We evaluate the proposed method on the cross-document RE task. We use the CodRED dataset~\cite{yao-etal-2021-codred}, which consists of 276 relation types derived from English Wikipedia and Wikidata. Documents in CodRED contain an average of 4,939 tokens.
Given our focus on evaluating cross-document reasoning, we conduct experiments in a closed setting where text paths are provided. We adopt End-to-end~\cite{yao-etal-2021-codred} and ECRIM~\cite{wang-etal-2022-entity} for the RE modules. We follow the settings provided by each model. We consider both BERT-base and RoBERTa-large~\cite{roberta} models as the embedding backbone. 

Regarding the REIC module, we use a fixed BERT-base~\cite{devlin-etal-2019-bert} model as the embedding backbone. We employ the AdamW~\cite{loshchilov2018decoupled} optimizer with a learning rate of 3e-3. Sentence embeddings are set to a dimension of 768. We use a 1-layer LSTM with a hidden vector dimension of 512. The policy network comprises a 2-layer feed-forward neural network with a hidden dimension of 512.

We use Snippet~\cite{yao-etal-2021-codred} and ECRIM-IC~\cite{wang-etal-2022-entity}, which are accessible implementations, as baselines for comparing the input construction modules. Snippet uses tokens near the target entity as input, while ECRIM-IC conducts heuristic sentence filtering based on the bridge entity. In line with previous work~\cite{yao-etal-2021-codred}, we use the AUC, F1, Precision@500 (P@500), and Precision@1000 (P@1000) as evaluation metrics. Further settings are specified in \cref{sec:app_exp_set}.

\paragraph{Reward Function}
We set the reward function to be proportional to the prediction score of the given relation, but different for each RE module due to the difference in their loss structures. For the End-to-End RE module, we set the reward function by the difference between the given relation score and the highest score excluding the relation: $R=\lambda_r ({\hat{y}_r - \max_{i\neq r} \hat{y}_i})/{\hat{y}_r}$. We set $\lambda_r=10$ for $r\neq \text{N/A}$ and $\lambda_r=1$ for $r=\text{N/A}$, and clip rewards below 0. For the ECRIM RE module, we set the reward function by the difference between the given relation score and threshold $\theta$ in the RE loss: $R=\lambda_r (\hat{y}_r - \theta)$, where $\lambda_r$ is the same as before. 

\subsection{Quantitative Results}

\cref{tab:main} presents the performance results of different input construction modules with various RE modules. We run experiments with all combinations of RE module backbones, their structures, and input construction module types. As shown in \cref{tab:main}, REIC demonstrates the best performance across all scenarios. In particular, the model combining the REIC module with ECRIM RE module using the RoBERTa backbone exhibits state-of-the-art performance on the development set with an AUC of 66.41 and F1 of 63.47, and on the test set with an AUC of 65.88 and F1 of 65.02.

Our proposed approach outperforms previous methods: Snippet + End-to-end~\cite{yao-etal-2021-codred}, with an increase of 1.97 in AUC and 1.34 in F1; and ECRIM-IC + ECRIM~\cite{wang-etal-2022-entity}, with an increase of 1.54 in AUC and 2.34 in F1, on the development set when using the BERT backbone. In the case of the input construction module of ECRIM, due to its heuristic nature, its performance even decreased compared to the Snippet approach on the End-to-end backbone that does not leverage bridge entity information (a decrease of 2.45/2.27 in AUC for BERT/RoBERTa backbones, respectively). Conversely, our method consistently improves performance in all scenarios because the selection module is trained based on the rewards received from each RE module.

\subsection{Analysis of Input Construction}

\begin{figure}[t]
    \centering
    \includegraphics[width=\linewidth]{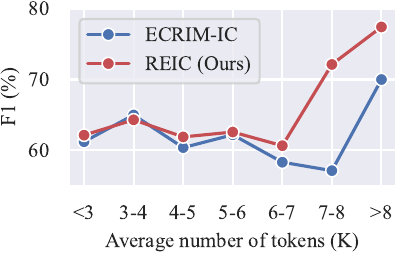}
    \caption{Experimental results on F1s based on the average number of tokens in a bag of text paths with ECRIM (BERT) RE module.}
    \label{fig:num_token}
\end{figure}

\paragraph{Effect Analysis for Document Length}
To analyze the effect of document length, we partition the CodRED development set based on the average number of tokens in the text paths for each entity pair. \cref{fig:num_token} presents the F1 score corresponding to the average number of tokens. We observe that our sentence selection method shows comparable performance in most cases, and especially greater effectiveness in cases with very long document lengths. This improvement is attributed to the dispersion of important sentences in longer documents, which allows our learning-based approach to select relevant sentences more effectively.

\begin{figure}[t]
    \centering
    \includegraphics[width=\linewidth]{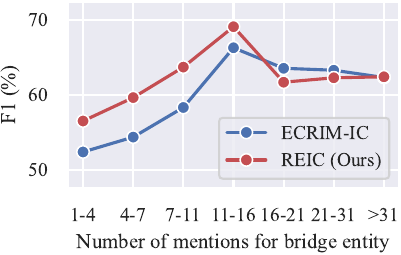}
    \caption{Experimental results on F1s based on the average number of bridge entities in a bag of text paths with ECRIM (BERT) RE module.}
    \label{fig:num_bridge}
\end{figure}

\paragraph{Effect Analysis for Number of Bridge Entities in Text Paths}
To analyze the effect of the number of bridge entities in text paths, we partition the CodRED development set based on the number of mentions of bridge entities in the text paths for each entity pair. \cref{fig:num_bridge} illustrates the variation in F1 scores according to the average number of mentions of bridge entities. Our model exhibits superior performance to ECRIM-IC, a heuristic construction method based on bridge entities, until the number of mentions reaches 16, which accounts for 64\% of the entity pairs in the dataset. However, we observe that the performance of REIC is slightly lower than ECRIM-IC when the number of mentions is very high. These results suggest that while ECRIM-IC struggles to distinguish important sentences when the number of mentions is insufficient; REIC, which learns directly from sentences, is more effective. However, when the number of bridge entities is large enough, the heuristic method based on bridge entities is slightly better. Therefore, we believe that incorporating bridge entity information into the reward function of our method can provide synergistic benefits, which we identify as a topic for future work.

\begin{table}
    \centering
    \adjustbox{max width=\linewidth}{%
        \begin{tabular}{lccc}
            \toprule
            Method & N/A Bag & Positive Bag \\
            \midrule
            ECRIM-IC & 5.01 & 9.87 \\
            REIC & 1.58 & 4.91 \\
            \bottomrule
        \end{tabular}
        }
        \captionof{table}{Average number of bridge entity mentions in selected sentences with ECRIM (BERT) RE module.} 
        \label{tab:avg_entity}
\end{table}

\begin{figure}[t]
    \centering
    \begin{subfigure}[t]{0.48\linewidth}
        \centering
        \includegraphics[width=\linewidth]{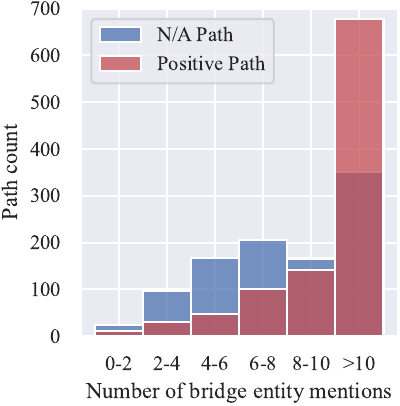}
        \caption{ECRIM-IC}
        \label{fig:hist_entity_ecrim}
    \end{subfigure}
    \hfill
    \begin{subfigure}[t]{0.48\linewidth}
        \centering
        \includegraphics[width=\linewidth]{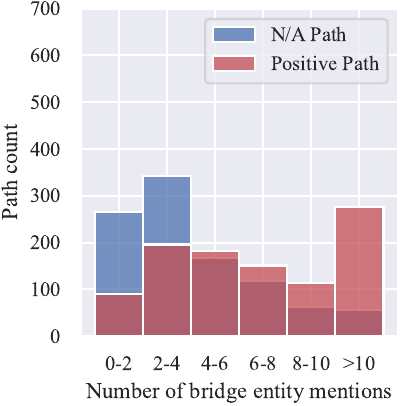}
        \caption{REIC (Ours)}
        \label{fig:hist_entity_reis}
    \end{subfigure}
    \caption{Histogram of the number of bridge entity mentions in sentences selected by (a) ECRIM-IC and (b) REIC for each document among entity pairs with positive relations, with ECRIM (BERT) RE module.}
    \label{fig:hist_entity}
\end{figure}

\begin{figure}[t]
    \centering
    \includegraphics[width=\linewidth]{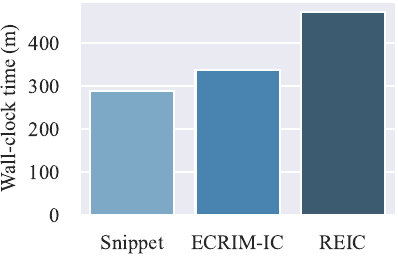}
    \caption{Comparison of training wall-clock time (minutes) per epoch with End-to-end (BERT) RE module.}
    \label{fig:wallclock}
\end{figure}

\begin{table}[t]
    \centering
    \adjustbox{max width=\linewidth}{%
    \begin{tabular}{lcccc}
        \toprule
        Method & AUC & F1 & P@500 & P@1000    \\
        \midrule
        Snippet & 53.83 & 55.17 & 67.80     & 54.80    \\
        ECRIM-IC       & 51.38 & 52.51 & 66.60     & 52.10    \\
        \midrule
        One-step & 54.88 & 56.32 & 72.00 & 56.00 \\
        $\lambda_r=1 \forall r$ & 54.65 & 56.05 & 70.00 & 56.50 \\
        \midrule
        REIC (Ours) & 55.80 & 56.51 & 71.20 & 56.70 \\
        \bottomrule
    \end{tabular}
    }
    \caption{Ablation studies on the CodRED Dev set with End-to-end (BERT) RE module.}
    \label{tab:abl}
\end{table}

\paragraph{Analysis for Number of Bridge Entities in Selected Sentences}

For the analysis of selected sentences, we examine the number of mentions of bridge entities in sentences selected by ECRIM-IC and REIC. \cref{tab:avg_entity} displays the average number of bridge entity mentions for each sentence selection module based on the relation type of entity pairs. Both modules extract more bridge entities from text paths of entity pairs with positive relations. ECRIM-IC extracts more bridge entities in all cases because it extracts sentences associated with bridge entities. Although our model is trained based on prediction scores without directly considering the number of bridge entities, it still extracts more bridge entities in positive relations. This observation highlights the importance of bridge entities in cross-document RE and suggests that our model effectively learns these bridge entities.

For entity pairs with positive relations, paths can either represent positive relations (positive paths) or not express the relation (N/A paths). We conduct an analysis of the number of bridge entities for each of these path types. \cref{fig:hist_entity} plots the number of bridge entity mentions for each sentence selection module in histograms. ECRIM-IC tends to extract more bridge entities for all path types. Conversely, REIC extracts fewer bridge entities in N/A paths. We interpret this phenomenon as a preference to extract fewer unnecessary or ambiguous entities from N/A paths, thereby focusing on extracting meaningful entities to infer relations.

\paragraph{Runtime Analysis}

\cref{fig:wallclock} compares the training wall-clock time per epoch for each input construction method. Unlike other methods, REIC introduces the sentence selection module, which takes more time. REIC takes about 1.40 times as long as the ECRIM-IC training time. Additionally, we observed that REIC takes about 1.75 times as long as the ECRIM-IC inference time. From these observations, the introduction of the REIC module leads to an increased time consumption. Currently, the selection process operates at the sentence level, involving the extraction of sentence-level embeddings and the computation of selection scores. However, we believe that extending this process to larger structural units, such as paragraphs or paths, could potentially reduce the number of selections and thus the overall time required.

\begin{figure}[t]
    \centering
    \includegraphics[width=\linewidth]{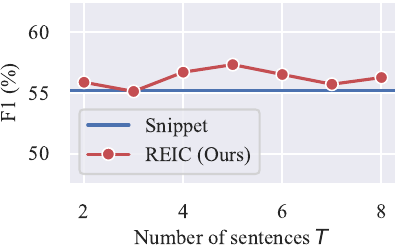}
    \caption{Ablation study for the number of selected sentences with End-to-end (BERT) RE module.}
    \label{fig:num_sen}
\end{figure}

\subsection{Ablation Studies}
\label{subsec:abl}

\paragraph{Multi-step vs. One-step Selection}
We employ a strategy of selecting one sentence at a time over $T$ multi-steps in the sentence selection process, inspired by \citet{Man_Ngo_Van_Nguyen_2022}. We explore the option of selecting $T$ sentences all at once in a one-step manner. In this case, we sample a sentence $T$ times based on the selection probability obtained in the first iteration. The evaluation time of the multi-step and one-step selections is almost the same because the policy network and the LSTM are much shallower compared to other networks such as BERT. As shown in the `One-step' row of \cref{tab:abl}, we observe an increase in performance of 0.92 in AUC with multi-step selection. We attribute this improvement to the LSTM providing information about the currently selected sentence, thus helping to select better sentences at each iteration.

\begin{table*}[t]
    \centering
    \adjustbox{max width=\linewidth}{%
    \begin{tabularx}{\textwidth}{sX}
        \toprule
        (\texttt{h,t,r}) & Selected sentences from REIC\\
        \midrule
        \small (\texttt{Chaosium}, \texttt{tabletop role-playing game}, product or material produced) & \small In 1974, “\textbf{Dungeons \& Dragons}” brought his interest to role-playing games. He became a full-time staff member at \texttt{Chaosium}. \newline A tabletop role-playing game (or \texttt{pen-and-paper role-playing game}) is a form of role-playing game (RPG) in which the participants describe their characters' actions through speech. ... \textcolor{purple}{Most games follow the pattern established by the first published role-playing game, “\textbf{Dungeons \& Dragons}”.}  \\
        \bottomrule
    \end{tabularx}
    }
    \caption{Sentences selected by REIC for the text path (`Sandy Petersen', `Tabletop role-playing game') containing the triplet (\texttt{Chaosium}, \texttt{tabletop role-playing game}, product or material produced). The first and second paragraphs are from the head and tail document, respectively. \textbf{Bold} font denotes the bridge entity, and \texttt{typewriter font} indicates the target entity. A sentence not found in both Snippet and ECRIM-IC is colored in \textcolor{purple}{purple}.}
    \label{tab:case}
\end{table*}

\paragraph{Reward Hyperparameter $\lambda_r$}
We set the reward hyperparameter to $\lambda_r=10$ for the positive relations to ensure that paths with positive relations receive higher rewards. This adjustment aims to address the problem that N/A paths occur about 14 times as often as positive paths, as reported in \citet{yao-etal-2021-codred}. To assess the effectiveness of this adjustment, we train a model with $\lambda_r$ set to $1$ for all relations. As shown in the `$\lambda_r=1 \forall r$' row of \cref{tab:abl}, we observe a performance improvement with REIC when assigning larger rewards to positive relations compared to when rewards are equally distributed. This suggests the necessity of adjusting rewards to account for the imbalance between positive and N/A relations, which is common in RE datasets. Moreover, considering potential imbalances among positive relations, further adjustments to reward hyperparameters may be necessary, which we identify as a direction for future work.

\paragraph{Number of Selected Sentences $T$}
In \cref{fig:num_sen}, we conduct an ablation study on the hyperparameter $T$, representing the number of selected sentences. We observe consistently better performance compared to the Snippet across different numbers of sentences. Increasing the number of sentences tends to improve performance up to a certain level, as it provides more information to the RE module. However, including more sentences than necessary may introduce noisy information, potentially affecting the ability to effectively distinguish relations.

\subsection{Case Study}
\label{sec:case}


\cref{tab:case} presents a case study of the key sentences selected by REIC to infer the entity relation, where the sentence missed by the baselines is highlighted in color. This sentence is crucial for inferring the relation between the entity pair. We also include the complete sets of sentences extracted by each input construction method in \cref{tab:app_case_snippet,tab:app_case_ecrim,tab:app_case_ours} of \cref{subsec:app_case}. Although ECRIM-IC extracts more bridge entities, REIC selects important sentences necessary to elucidate entity relations.

\section{Conclusion}
\label{sec:conc}

We present REward-based Input Construction (REIC), a learning-based sentence selection module tailored for cross-document RE. We tackle the limitation of document embedding extraction from long documents. REIC selects important sentences based on relational evidence, enabling the RE module to effectively infer relations. Given the unavailability of supervision for evidence sentences, we use RL to train REIC. We show the superiority of our method over heuristic methods. This highlights the potential of learning-based approaches to improve the performance of cross-document RE.

\section*{Limitations}

Limitations of our method include that the introduction of a network for input construction increases the model runtime by a factor of 1.40 for training and 1.75 for inference in the ECRIM-IC runtime. To alleviate this, we precompute and store sentence embeddings from BERT to avoid repetitive BERT evaluations during training. One possible solution is to apply model compression methods, such as pruning and weight factorization, which have been used in many NLP studies~\cite{Lan2020ALBERT,gordon-etal-2020-compressing,wang-etal-2020-structured}.

\section*{Acknowledgements}

This research was supported by Research and Development on Key Supply Network Identification, Threat Analyses, Alternative Recommendation from Natural Language Data (NRF) funded by the Ministry of Education (2021R1A2C200981612).

\bibliography{main}

\clearpage
\appendix

\section{Further Review of Related Work}
\label{sec:app_rel}

Early relation extraction models were proposed to predict relationships between entities within a single sentence~\cite{zeng2014relation,wang2016relation,zhang2017position,zhu-etal-2019-graph}. Some research on sentence-level relation extraction introduced a novel convolutional neural network (CNN) architecture to extract sentence-level features from word tokens \cite{zeng2014relation,wang2016relation}. Additionally, \citet{zhang2017position} not only presented an LSTM sequence model with entity position-aware attention, but also proposed TACRED, a large supervised dataset for sentence-level RE, serving as a benchmark dataset. GP-GNNs \cite{zhu-etal-2019-graph} propose a novel graph neural network where its parameters are generated through a propagated message passing module taking sentences as input.

\citet{yao-etal-2019-docred} introduced DocRED, a comprehensive dataset with human-annotated document-level relation extraction, aiming to advance from sentence-level to document-level relation extraction, offering a wealth of relation facts. 
Several methods for document-level RE utilize document-level graphs constructed from entities or mentions, employing graph neural networks (GNN) for path reasoning to infer relations between entities \cite{sahu-etal-2019-inter, nan-etal-2020-reasoning, zeng2020double}. On the other hand, many document-level RE methods have emerged that do not rely on graph structures. \citet{huang-etal-2021-three} presents a straightforward method to heuristically select evidence sentences. SSAN \cite{xu2021entity} integrates structural dependencies into self-attention, enhancing attention flow for relation extraction. Recent challenges for document-level RE include high memory usage and limited annotations. DREEAM \cite{ma-etal-2023-dreeam} and S2ynRE \cite{xu-etal-2023-s2ynre} offer a memory-efficient solution and self-training to generate evidence and training data.

\section{Additional Experimental Settings}
\label{sec:app_exp_set}

\subsection{Datasets}
\label{subsec:app_dataset}
The CodRED dataset, sourced from its official Github repository, comprises relation triplets, evidence, and documents, licensed under MIT. CodRED also includes processed Wikidata (CC0 license) and Wikipedia (licensed under CC BY-SA and GFDL). We processed the raw data as per repository instructions, transforming and storing documents in a Redis server for downstream RE methods. In particular, for efficiency, we stored BERT embeddings for each sentence of every document in the server environment for operational efficiency. CodRED comprises 276 relations, and its statistics are presented in \cref{tab:codred_stat}.

\begin{table}
    \centering
    \adjustbox{max width=\linewidth}{%
        \begin{tabular}{lccc}
            \toprule
            Split & Triplets & Text paths (mean) \\
            \midrule
            Train & 19,461 & 129,548 (6.6) \\
            Dev & 5,568 & 40,740 (7.3) \\
            Test & 5,535 & 40,524 (7.3) \\
            \bottomrule
        \end{tabular}
        }
        \caption{The statistics of CodRED dataset.}
        \label{tab:codred_stat}
\end{table}

\subsection{Implementation Details}
\label{subsec:app_imp}
\subsubsection{Model Parameters}
\label{subsubsec:app_imp_params}
Table \ref{tab:params} displays the number of parameters for different models. Each row represents a RE module's backbone, structure, an input construction module and the corresponding number of parameters. For BERT, the parameter count is 108,310,272, while for RoBERTa, it is 355,462,144. Our REIC model introduces a selector module, increasing the parameter count by 3,281,921 for each RE module.

\begin{table}[h]
    \centering
    \resizebox{\linewidth}{!}{%
    \begin{tabular}{|c|c|c|c|}
        \hline
        \textbf{Backbone} & \textbf{RE Module} & \textbf{IC Module} & \textbf{\# of Parameters} \\
        \hline
        \multirow{2}{*}{BERT} & \multirow{2}{*}{End-to-end} & Snippet & 108,523,285 \\
        \cline{3-4}
        & & REIC (Ours) & 111,805,206 \\
        \hline
        \multirow{2}{*}{RoBERTa} & \multirow{2}{*}{End-to-end} & Snippet & 355,746,069 \\
        \cline{3-4}
        & & REIC (Ours) & 358,028,990 \\
        \hline
        \multirow{2}{*}{BERT} & \multirow{2}{*}{ECRIM} & ECRIM-IC & 122,225,509 \\
        \cline{3-4}
        & & REIC (Ours) & 125,844,193 \\
        \hline
        \multirow{2}{*}{RoBERTa} & \multirow{2}{*}{ECRIM} & ECRIM-IC & 375,111,397 \\
        \cline{3-4}
        & & REIC (Ours) & 388,063,055 \\
        \hline
    \end{tabular}
    }
    \caption{Number of parameters for each model. RE is short for relation extraction and IC is short for input construction.}
    \label{tab:params}
\end{table}

\subsubsection{Code Implementation}
\label{subsubsec:app_imp_baseline}
During the implementation process, we utilized the official codes of End-to-end\footnote{ \url{https://github.com/thunlp/CodRED}}~\cite{yao-etal-2021-codred} and ECRIM\footnote{ \url{https://github.com/MakiseKuurisu/ecrim}}~\cite{wang-etal-2022-entity}. We trained both the end-to-end and ECRIM models using a learning rate of 3e-5. The training process involved running 2 epochs for the End-to-end and 10 epochs for the ECRIM. One change is to set the batch size to 4 and the gradient accumulation step to 4 for ECRIM.

End-to-end model underwent the modification during implementation that had a significant impact on performance. In the original code, the entity marking token `[UNUSED@]', where `@' is a natural number, was utilized. However, we found that this token was functionally meaningless and replaced it with `[unused@]'. Then, the performance of the modification, as shown in the row of `Snippet + End-to-end' in \cref{tab:main}, exhibited a significant improvement compared to the results reported in \citet{yao-etal-2021-codred}.

Our REIC model implementation is also based on the official End-to-end and ECRIM codebases. Additionally, we incorporated the sentence selection module from SCS-EERE\footnote{\url{https://github.com/hieumdt/SCS-EERE}}~\cite{Man_Ngo_Van_Nguyen_2022} as a reference. If the selected $T$ sentences exceed the token limit, we use only the first 512 tokens in the order they were chosen. Our REIC model was trained according to the settings specified for each RE structure, and detailed training configurations for our REIC are provided in \cref{subsec:exp_setting}. We utilized a single NVIDIA A100 80GB model for training both the baseline and our model. 
We performed a single run for each of the various RE backbones and structures to produce the results.



\begin{table}[t]
    \centering
    \adjustbox{max width=\linewidth}{%
    \begin{tabular}{cc}
        \toprule
        Average number of tokens (K) & Number of cases   \\
        \midrule
        < 3 & 743 \\
        3-4 & 1071 \\
        4-5 & 1227 \\
        5-6 & 1013 \\
        6-7 & 655 \\
        7-8 & 382 \\
        > 8 & 477 \\
        \bottomrule
    \end{tabular}
    }
    \caption{Number of cases in each partition by average number of tokens, which is related to \cref{fig:num_token}.}
    \label{tab:partition_4}
\end{table}

\begin{table}[t]
    \centering
    \adjustbox{max width=\linewidth}{%
    \begin{tabular}{cc}
        \toprule
        Number of mentions for bridge entity & Number of cases   \\
        \midrule
        1-4 & 665 \\
        4-7 & 889 \\
        7-11 & 1085 \\
        11-16& 943 \\
        16-21 & 587 \\
        21-31 & 590 \\
        > 31 & 809 \\
        \bottomrule
    \end{tabular}
    }
    \caption{Number of cases in each partition by average number of bridge entity, which is related to \cref{fig:num_bridge}.}
    \label{tab:partition_5}
\end{table}

\subsection{Details of Input Construction Analysis}

\cref{tab:partition_4,tab:partition_5} provide the number of bags in each partition for each analysis.

\section{Additional Experimental Results}
\label{sec:app_exp_res}

\begin{figure}[t]
    \centering
    \includegraphics[width=0.8\linewidth]{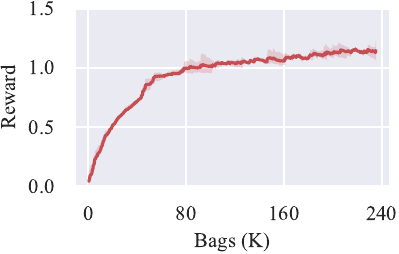}
    \caption{Reward curve during training with End-to-end (BERT) RE module. We use the exponential moving average with factor of 0.99. We ran the experiment three times. Solid line represents the mean, and the shaded region illustrates the minimum and maximum values.}
    \label{fig:reward_plot}
\end{figure}

\subsection{Reward Analysis}
\label{subsec:app_reward}

\cref{fig:reward_plot} shows the selection reward curve during training. As training progresses, the reward gradually increases. After a certain initial step, the reward stabilizes at 1, which is a reward value for correct predictions of the N/A relations. This is probably influenced by the large proportion of N/A relations in the dataset and the model correctly predicts the N/A relations. Subsequently, in later stages, the reward surpasses 1, indicating that the model is increasingly focused on correctly predicting positive relations.

\subsection{Case Study}
\label{subsec:app_case}

\cref{tab:app_case_snippet,tab:app_case_ecrim,tab:app_case_ours} provide the complete set of sentences selected by each input construction module for the case study analyzed in \cref{sec:case}. 

\begin{table*}[t]
    \centering
    \adjustbox{max width=0.9\linewidth}{%
    \begin{tabularx}{\textwidth}{cX}
        \toprule
        Source & Selected sentences from Snippet \\
        \midrule
        Head & “Sandy Petersen” (born September 16, 1955) is an American game designer. Petersen was born in St. Louis, Missouri and attended University of California, Berkeley, majoring in zoology. He is a well-known fan of H. P. Lovecraft, whose work he first encountered in a \textbf{World War II} Armed Services Edition of “The Dunwich Horror and other Weird Tales” found in his father's library. In 1974, “\textbf{Dungeons \& Dragons}” brought his interest to role-playing games. He became a full-time staff member at \texttt{Chaosium}. His interest for role-playing games and H. P. Lovecraft were fused when he became principal author of Chaosium's game. “\textbf{Call of Cthulhu}”, published 1981, and many scenarios and background pieces thereafter. While working for Chaosium he co-authored the third edition of “\textbf{RuneQuest}”, for which he also co-wrote the critically acclaimed “Trollpak” and a number of other Gloranthan supplements. He authored several critically acclaimed “RuneQuest” supplements for Avalon Hill and \\
        \midrule
        Tail & A tabletop role-playing game (or \texttt{pen-and-paper role-playing game}) is a form of role-playing game (RPG) in which the participants describe their characters' actions through speech. Participants determine the actions of their characters based on their characterization, and the actions succeed or fail according to a set formal system of rules and guidelines. Within the rules, players have the freedom to improvise; their choices shape the direction and outcome of the game. Unlike other types of role-playing game, tabletop RPGs are often conducted like radio drama: only the spoken component of a role is acted. This acting is not always literal, and players do not always speak exclusively in-character. Instead, players act out their role by deciding and describing what actions their characters will take within the rules of the game. In most games, a specially designated player called the game master (GM) — also known as the Dungeon Master (DM) in “\textbf{Dungeons \& Dragons}”, Referee for all Game Designers' Workshop games, or Storyteller for the Storytelling System — creates a setting in which each player plays the role of a single character. The GM describes the game world and \\
        \bottomrule
    \end{tabularx}
    }
    \caption{Example of the selected sentences from Snippet. The first row represents the sentences selected from the head document, and the second row represents the sentences selected from the tail document. Text in typewriter font indicates the target entity, while bold text indicates the bridge entity.}
    \label{tab:app_case_snippet}
\end{table*}

\begin{table*}[t]
    \centering
    \adjustbox{max width=0.9\linewidth}{%
    \begin{tabularx}{\textwidth}{cX}
        \toprule
        Source & Selected sentences from ECRIM-IC \\
        \midrule
        Head & Lovecraft, whose work he first encountered in a \textbf{World War II} Armed Services Edition of “The Dunwich Horror and other Weird Tales” found in his father' s library. In 1974, “\textbf{Dungeons \& Dragons}” brought his interest to role-playing games. He became a full-time staff member at \texttt{Chaosium}. His interest for role-playing games and H. P. Lovecraft were fused when he became principal author of Chaosium' s game “\textbf{Call of Cthulhu}”, published 1981, \\
        \midrule
        Tail & A tabletop role - playing game (or \texttt{pen-and-paper role-playing game}) is a form of role-playing game (RPG) in which the participants describe their characters' actions through speech. Participants determine the actions of their characters based on their characterization, In most games, a specially designated player called the game master (GM) — also known as the Dungeon Master (DM) in “\textbf{Dungeons \& Dragons}”, Referee for all Game Designers' Workshop games, Due to the game' s success, the term “\textbf{Dungeons \& Dragons}” has sometimes been used as a generic term for fantasy role-playing games. TSR undertook legal action to prevent its trademark from becoming generic. The “d20 system”, based on the third edition of “\textbf{Dungeons \& Dragons}”, was used in many modern or science fiction game settings such as “Spycraft” and the “Star Wars Roleplaying Game”. Usually a campaign setting is designed for a specific game (such as the “Forgotten Realms” setting for “\textbf{Dungeons \& Dragons}”) or a specific genre of game (such as Medieval fantasy, \textbf{World War II}, or outer space / science fiction adventure ). horror formed the baseline of the “World of Darkness” and “\textbf{Call of Cthulhu}” while “Spycraft” was based in modern-day spy thriller-oriented settings. \\
        \bottomrule
    \end{tabularx}
    }
    \caption{Example of the selected sentences from ECRIM-IC. The first row represents the sentences selected from the head document, and the second row represents the sentences selected from the tail document. Text in typewriter font indicates the target entity, while bold text indicates the bridge entity.}
    \label{tab:app_case_ecrim}
\end{table*}

\begin{table*}[t]
    \centering
    \adjustbox{max width=0.9\linewidth}{%
    \begin{tabularx}{\textwidth}{cX}
        \toprule
        Source & Selected sentences from REIC \\
        \midrule
        Head & Petersen was born in St. Louis, Missouri and attended University of California, Berkeley, majoring in zoology. He is a well-known fan of H. P. Lovecraft, whose work he first encountered in a \textbf{World War II} Armed Services Edition of “The Dunwich Horror and other Weird Tales” found in his father' s library. In 1974, “\textbf{Dungeons \& Dragons}” brought his interest to role-playing games. He became a full-time staff member at \texttt{Chaosium}. His interest for role-playing games and H. P. Lovecraft were fused when he became principal author of Chaosium' s game “\textbf{Call of Cthulhu}”, published 1981, and many scenarios and background pieces thereafter. While working for Chaosium he co-authored the third edition of “\textbf{RuneQuest}”, for which he also co-wrote the critically acclaimed “Trollpak” and a number of other Gloranthan supplements. and is a frequent guest at conventions where he usually runs a freeform game of his own devising, and / or helps to run someone else' s game. He worked some \\
        \midrule
        Tail & A tabletop role-playing game (or \texttt{pen-and-paper role-playing game}) is a form of role-playing game (RPG) in which the participants describe their characters' actions through speech. Participants determine the actions of their characters based on their characterization, Most games follow the pattern established by the first published role-playing game, “\textbf{Dungeons \& Dragons}”. Participants usually conduct the game as a small social gathering. One participant, called the Dungeon Master (DM) in “\textbf{Dungeons and Dragons}”, more commonly called the game master or GM, purchases or prepares a set of rules and a fictional setting in which players can act out the roles of their characters. from a single brief session (usually completed in a few hours) to a series of repeated sessions that may continue for years with an evolving cast of players and characters. Play is often episodic and mission-centric, with a series of challenges culminating in a final puzzle or enemy that must be overcome. Gygax expected to sell about 50, 000 copies total to a strictly hobbyist market. After establishing itself in boutique stores, it developed a strong, lasting fan base \\
        \bottomrule
    \end{tabularx}
    }
    \caption{Example of the selected sentences from REIC (Ours). The first row represents the sentences selected from the head document, and the second row represents the sentences selected from the tail document. Text in typewriter font indicates the target entity, while bold text indicates the bridge entity.}
    \label{tab:app_case_ours}
\end{table*}

\section{Potential Risks}
\label{sec:app_pot_risk}

Our method may select incorrect or irrelevant sentences, resulting in inaccurate relation extraction. This risk could occur if there are limitations in the training data. In addition, there could be cases of noisy or harmful labels in the training data, which could affect the extraction of harmful sentences.

\end{document}